\pdfoutput=1

\documentclass[11pt]{article}

\usepackage{acl}
\usepackage{hyperref}
\usepackage{times}
\usepackage{latexsym}
\usepackage{multirow}
\usepackage{booktabs}
\usepackage{microtype}
\usepackage{multirow}
\usepackage{latexsym}
\usepackage{booktabs}
\usepackage{courier}
\usepackage{amsmath}
\usepackage{booktabs} 
\usepackage{siunitx} 
\usepackage{subfig}
\usepackage{placeins}
\usepackage{tikz}
\usepackage{cleveref}

\usepackage[T1]{fontenc}

\usepackage[utf8]{inputenc}

\usepackage{microtype}

\usepackage{inconsolata}

\usepackage{authblk}
\usepackage{amssymb}
\usepackage{hologo}
\title{FZI-WIM at SemEval-2024 Task 2: Self-Consistent CoT for Complex NLI in Biomedical Domain}

\author[1, 2]{Jin Liu}

\author[1]{Steffen Thoma}

\affil[1]{FZI Research Center for Information Technology, Karlsruhe, Germany}
\affil[2]{Karlsruhe Institute of Technology, Karlsruhe, Germany}
\affil[ ]{\text{\{jin.liu, thoma\}@fzi.de}}

\begin{document}
\maketitle
\begin{abstract}
This paper describes the inference system of FZI-WIM at the SemEval-2024 Task 2: Safe Biomedical Natural Language Inference for Clinical Trials. Our system utilizes the chain of thought (CoT) paradigm to tackle this complex reasoning problem and further improves the CoT performance with self-consistency. Instead of greedy decoding, we sample multiple reasoning chains with the same prompt and make the final verification with majority voting. The self-consistent CoT system achieves a baseline F1 score of 0.80 (1st), faithfulness score of 0.90 (3rd), and consistency score of 0.73 (12th). We release the code and data publicly\footnote{\url{https://github.com/jens5588/FZI-WIM-NLI4CT}}.
\end{abstract}

\section{Introduction}
The Safe Biomedical Natural Language Inference for Clinical Trials (NLI4CT) task aims to investigate the consistency and faithfulness of natural language inference (NLI) models in clinical settings\cite{jullien-etal-2024-semeval}. NLI is a typical natural language task requiring natural language reasoning~\cite{yu2023natural}. Fine-tuned BERT-based~\cite{devlin-etal-2019-bert} discriminative models have been widely applied to solve NLI problems~\cite{liu2020roberta,he2021deberta}. Studies show increasing reasoning capabilities of large language models (LLMs), both in proprietary \cite{gpt3, openai2023gpt4} and open-source LLMs \cite{touvron2023llama,touvron2023llama2,jiang2024mixtral}. However, problems of inconsistency and unfaithfulness still occur with LLMs\cite{golovneva2023roscoe,turpin2023language}. Compared to other domains, medical applications have much higher standards regarding safety and trustfulness, so inconsistencies and unfaithfulness limit AI applications in the medical domain.

Chain of Thought (CoT) has been proposed to elicit the reasoning capabilities of LLMs \cite{cot}. Based on the CoT, further concepts like Tree of Thought \cite{tot}, ReAct \cite{yao2023react}, Self-Consistency \cite{wang2023selfconsistency}, and so on have been proposed to improve the performance of CoT further. One common characteristic of the frameworks mentioned above is the explicit generation of reasoning chains. Since only verification labels are provided in the NLI4CT training dataset, we utilize GPT-4 to generate reasoning chains. With the distilled knowledge from GPT-4, we further instruction-tune an open-source LLM with low-rank adaption (LoRA) \cite{hu2022lora} for claim verification with CoT. Our system follows the self-consistency concept by generating multiple CoT reasoning chains and verifying with majority voting. 

We summarize our major findings regarding this task as follows:
\begin{itemize}
        \item LoRA instruction-tuning can bring domain-specific (biomedical) knowledge and reasoning capabilities to LLMs.
        \item Our instruction-tuned LLM tends to contradict the statement if the information is only contained in the statement, not in the premise, even if the information is factually correct. 
        \item CoT reasoning gains significant performance improvement regarding faithfulness compared to the label-only prediction.
        
        \item Compared to the greedy CoT, self-consistent CoT with majority voting has a performance improvement of 1.31 (baseline F1), 0.75 (consistency score), and 0.69 (faithfulness score) percentage points. Performance improvement is limited for the binary classification problem.
    \end{itemize}

\section{Background}
\begin{figure}
\includegraphics[scale=0.39]{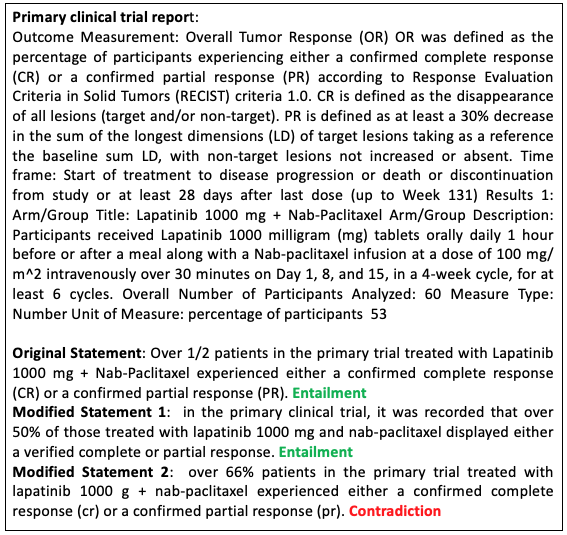}
\caption{A data example. With the same clinical report, semantic-preserving and semantic-altering interventions on the original statement are used  to evaluate the consistency and faithfulness of the verification system.}
\label{fig:example}
\end{figure}

NLI4CT contains one text entailment task, namely infer the relationship between a premise and a statement as either entailment or contradiction~\cite{jullien-etal-2024-semeval}. The premises in the NLI4CT dataset are collected from publicly available breast cancer clinical trial reports (CTRs), split into four sections: eligibility criteria, intervention, results, and adverse event~\cite{jullien-etal-2023-semeval}. The statements are sentences making claims about the information in the CTR premise, either about a single CTR or a comparison between 2 CTRs.  The numbers of training, validation, and test examples are 1700, 200, and 5500, respectively. In the test set, 500 examples are used as anchors. The other 5000 statements are created with interventions on these first 500 examples. \Cref{fig:example} shows an example with interventions on the statements. Using the same clinical report, two statements are modified, one semantics-preserving (modified statement 1) and the other semantics-altering (modified statement 2), based on the original statement.  The purpose of the interventions is to investigate the consistency and faithfulness of the inference.

For the text entailment task in the first iteration of NLI4CT (SemEval-2023 Task 7), most systems have fine-tuned discriminative transformer-based models \cite{jullien-etal-2023-semeval}.  \citet{kanakarajan-sankarasubbu-2023-saama} instruction-tuned Flan-T5 model and achieved the 2nd place for the text entailment task. However, the systems mentioned above only predicted the verification labels without the reasoning process. To achieve a trustworthy verification, our system not only predicts the label but also the reasoning chains. Since the training and validation datasets have only provided verification labels. We utilize GPT-4 to verify with reasoning chains for training and validation datasets. We further instruction-tune an open-source LLM with distilled reasoning chains from GPT-4. To address the inconsistency problem in reasoning chains, we sample multiple chains and then employ a majority voting approach to determine the final verification outcome.

\section{System overview}
\begin{figure*}
\includegraphics[scale=0.49]{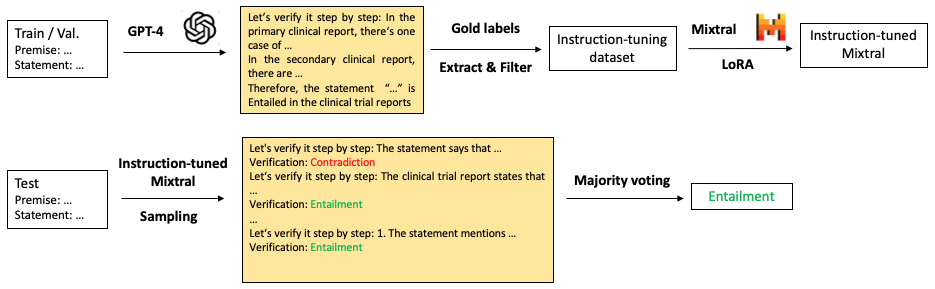}
\caption{Training and Inference pipeline of self-consistent CoT system.}
\label{fig:pipeline}
\end{figure*}
Figure~\ref{fig:pipeline} shows the pipelines of data creation, model training, and inference of our system. In the following, we describe each part.

\subsection{Knowledge Distillation}
The NLI4CT data only contains verification labels without rationales. Following the concept of knowledge distillation \cite{hinton2015distilling}, we leverage GPT-4 to generate rationales with CoT for the training and validation datasets. We further extract the verification labels of GPT-4 based on the CoT reasoning and filter out the examples for which the extracted labels are different from the gold labels.

\subsection{LoRA Instruction-tuning}
Parameter efficient fine-tuning (PEFT) gains popularity as the sizes of LLMs increase~\cite{pmlr-v97-houlsby19a,pfeiffer-etal-2020-adapterhub,li-liang-2021-prefix,hu2022lora}. Low-rank adaption (LoRA)~\cite{hu2022lora} is a PEFT approach to fine-tune LLMs. Studies show over-parametrized models reside on a low intrinsic dimension \cite{aghajanyan-etal-2021-intrinsic,Li2018MeasuringTI,hu2022lora}. The key assumption of LoRA is that the updates to the weights also have a low intrinsic rank during adaptation for downstream tasks. The parameter updates of a pre-trained weight matrix $W_{0} \in \mathbb{R}^{d\times k}$ can be represented as $W_{0}+\Delta W=W_{0}+BA$, where $B\in \mathbb{R}^{d\times r}$, $A\in \mathbb{R}^{r\times k}$ and $r\ll \min(d,k)$ \cite{hu2022lora}. 

Based on the concept of LoRA, given prompt $x$ and the target output $y=(y_{1},...,y_{m})$, the loss can be formulated as:
\begin{equation}
L=\sum_{i=1}^{m}-\log(p_{\theta}(\hat{y_{i}}=y_{i}|x, y_{1},...,y_{i-1})),
\end{equation}
where $\theta$ represents $W_{0}$, $B$, $A$ and only $B$ and $A$ are trainable.

\subsection{Self-Consistency}
Self-consistency for LLMs was proposed by \citet{wang2023selfconsistency} to replace the greedy decoding in CoT reasoning. The intuition behind self-consistency is that there are multiple ways to solve a complex problem. Another fact that supports the introduction of the self-consistency concept is that current LLMs still have difficulties with complex reasoning tasks and make mistakes in certain reasoning steps. With multiple reasoning chains, the model is less likely to make the same error and reach the same wrong answer. \citet{wang2023selfconsistency} show self-consistency boosts the performance of different LLMs with different sampling strategies, including temperature sampling, top-$k$ sampling, and nucleus sampling, on arithmetic, commonsense and symbolic reasoning tasks.

\section{Experimental setup}

In this section, we describe our experiment setup. We describe the CoT generation with GPT-4, the model training, and the inference setups.
\subsection{LoRA Instruction-tuning}

\noindent\textbf{Instruction Data Creation} Since GPT-4 has state-of-the-art reasoning capabilities, we instruct GPT-4 to verify the statements based on the CTR premise step-by-step for the training and validation datasets. The instruction prompt is shown in \Cref{sec:instruction_gpt4}. We then extract the verification labels of CoT rationales with an NLI model, namely bart-large-mnli~\cite{lewis-etal-2020-bart}, and keyword matching. We compare the extracted labels with gold labels and filter out examples where the verification is wrong. For 1700 training and 200 validation examples,  we achieve 1413 and 166 correctly verified examples with CoT, respectively, resulting in an accuracy of approximately 83\%. With these selected examples, we build an instruction-tuning dataset for fine-tuning an open-source LLM. We show an example of CoT instruction-tuning data in \Cref{sec:cot_instrcution_data_example}. For further comparison, we create another instruction dataset containing only verification labels without CoT for training and validation datasets. \Cref{sec:label_only_instrcution_data_example} shows one label-only instruction-tuning data example.

\noindent\textbf{Instruction-tuning} We select Mixtral-8x7B-Instruct~\cite{jiang2024mixtral} as our open-source LLM for instruction-tuning. The base model selection has considered reasoning capabilities and the number of model parameters. The configuration of LoRA is set as follows: r=8, alpha=32 and adaption of the attention weights of query ($W_{q}$), key ($W_{k}$), value ($W_{v}$) and output ($W_{o}$). This setup leads to 6.8M trainable parameters, which corresponds to 0.0146\% of the base model size of 46.7B. We instruction-tune Mixtral-8x7-Instruct with the CoT instruction dataset and label-only instruction data separately for five epochs. Further implementation details are described in \Cref{appendix:instruction_tuning}.

\subsection{Inference}
\noindent\textbf{Self-Consistent Inference} To generate reasoning chains with the CoT instruction-tuned Mixtral model, we follow \citet{wang2023selfconsistency} using a temperature sampling with the parameters $T=0.7$ and $k=50$ in top-$k$. For each (premise, statement) pair in the test dataset, we begin with sampling 10 reasoning chains. We then extract the verification labels and apply majority voting to decide the final label. If the result for both labels is tied, we further generate reasoning chains with different seeds. The maximal number of generated reasoning chains is 25. We show the number of generated reasoning chains in \Cref{appendix:model_inference}.

\noindent\textbf{Greedy Inference} For comparison, we also apply the greedy decoding strategy to generate verification with label-only and CoT instruction-tuned models. With the greedy strategy, the models predict the next token with the highest probability without sampling\footnote{\url{https://huggingface.co/docs/transformers/generation_strategies}}.

\subsection{Evaluation Metrics} 
Three evaluation metrics are used for the task: baseline F1, faithfulness score, and consistency score. The F1 score is used to evaluate the performance of the control set without intervention, consisting of 500 samples. There are 5000 samples in the test dataset, created with interventions based on the samples in the control set. 

The consistency score evaluates whether the model predicts the same label when semantic-preserving interventions exist in the original statements~\cite{jullien-etal-2024-semeval}. For $N$ statements $x_{i}$ in the contrast set $C$, their corresponding original statements $y_{i}$ and model prediction function $f$, the consistency score is calculated with:
\begin{equation}
Consistency=\frac{1}{N}\sum_{1}^{N}1-\left|f(y_{i})-f(x_{i})\right|
\end{equation}
where $x_{i}\in C$: $\text{Label}(x_{i})=\text{Label}(y_{i})$.

The faithfulness score evaluates whether the model can change its prediction when semantic-altering interventions are present in the original statements~\cite{jullien-etal-2024-semeval}. For $N$ statements $x_{i}$ in the contrast set $C$, their corresponding original statements $y_{i}$ and model prediction function $f$, the faithfulness score is calculated with:
\begin{equation}
Faithfulness=\frac{1}{N}\sum_{1}^{N}\left|f(y_{i})-f(x_{i})\right| 
\end{equation}
where $x_{i}\in C$: $\text{Label}(x_{i})\neq\text{Label}(y_{i})$, and $f(y_{i})=\text{Label}(y_{i})$. 

\section{Evaluation}
In this section, we report the results of our self-consistent CoT system. For comparison, we also report the results of label-only verification and CoT-greedy. The comparison of three systems serves as an ablation study to analyze the performance improvement of extra steps, reasoning chain generation and self-consistent verification. \Cref{tab:overall_comparison} gives an overview of the three systems compared to the best scores in each category. Our self-consistent CoT system achieves 1st place regarding baseline F1, 3rd place regarding faithfulness, and 12th place regarding consistency. Next, we will analyze our system according to the three metrics separately.

\begin{table}[t]
\centering
\resizebox{\columnwidth}{!}{
\begin{tabular}{@{}lccc@{}}
\toprule
\textbf{Model} & \textbf{Base F1} & \textbf{Consistency} & \textbf{Faithfulness} \\
\midrule
Label-only Greedy & 0.7867 & \textbf{0.7364} & 0.8102 \\
CoT Greedy & 0.7869 & 0.7217 & 0.8970 \\
Self-Consistent CoT  & \textbf{0.8000} & 0.7292 & \textbf{0.9039} \\
\midrule
Best score & 0.80 &0.81 &0.95\\
\bottomrule
\end{tabular}%
}
\caption{Overview of three systems compared to the best scores of the task regarding each metric.}
\label{tab:overall_comparison}
\end{table}

\subsection{Baseline F1}
In the test dataset, there are 500 samples without interventions on the statements, which are used as the control set. \Cref{tab:f1_score} shows the F1 scores of three systems according to the sections in the CTRs. The Eligibility section has the lowest F1 scores compared to the other sections in CTRs. Several factors have increased the difficulties for eligibility verification statements. First, the premises, i.e., criteria for inclusion and exclusion for clinical trials, are much longer than those for other sections. LLMs cannot always extract all relevant information from very long contexts. The verification of statements regarding eligibility often requires multi-step reasoning capabilities. The lack of domain-specific knowledge and common sense can also lead to verification errors. For the sections Intervention, Adverse Events, and Results, the major error type is numerical reasoning. 

\begin{table}[t]
\centering
\resizebox{\columnwidth}{!}{
\begin{tabular}{@{}lcccc@{}}
\toprule
\textbf{Model} & \textbf{Eligibility} & \textbf{Intervention} & \textbf{Adverse Events} &\textbf{Results}\\
\midrule
Label-only Greedy   & 0.7482 & \textbf{0.8175} & 0.8131 &0.7679 \\
CoT Greedy          & 0.7132 & 0.7794 & \textbf{0.8667} & 0.7961 \\
Self-Consistent CoT & \textbf{0.7581} & 0.7770 & 0.8305 &\textbf{0.8485} \\

\bottomrule
\end{tabular}%
}
\caption{F1 scores of three systems according to the sections in CTRs.}
\label{tab:f1_score}
\end{table}

\subsection{Consistency}
Compared to the other two metrics, the consistency score of our self-consistent CoT system has the worst ranking. There are 4136 samples with semantic-preserving interventions, and they can be classified into five groups: paraphrase preserving, contradiction preserving, numerical paraphrase preserving, numerical contradiction preserving, and definitions preserving. \Cref{tab:consistency_comparison} summarizes the consistency scores for each system according to these categories. Compared to the other categories, self-consistent CoT and CoT greedy systems have worse performance regarding definitions and numerical paraphrase interventions. For the intervention regarding definitions, extra factual statements have been added to the original statements. A definition-intervention to the original statement in \Cref{fig:example}, e.g., is:

\noindent\textit{Over 1/2 patients in the primary trial treated with Lapatinib 1000 mg + Nab-Paclitaxel experienced either a confirmed complete response (CR) or a confirmed partial response (PR). bladder solitary fibrous tumor is a solitary fibrous tumor that arises from the bladder. most tumors are benign.}

\noindent Since the sentence \textit{bladder solitary fibrous ...} is irrelevant to the clinical report, our self-consistent CoT tends to verify it as Contradiction (7 Contradictions : 3 Entailments in 10 generated reasoning chains).

For the category of numerical paraphrasing, there are 224 statements modified from 90 original statements. For the original set, we have around 81\% accuracy (73 / 90), and for the modified set around 72\% (162 / 224). We classify the errors, in total 34, where original statements are verified correctly and modified statements are verified wrong, into three groups: the conversion of fractional numbers, decimals, and percentages; the conversion of time, e.g., months to weeks, months to days, etc.; the conversion of units, e.g., mm to cm, mg to micrograms, etc. Among 34 misclassified statements, the system can generate at least one completely correct reasoning chain out of 10 reasoning chains for 17 statements. This indicates that the model has the knowledge for these conversions. However, our instruction-tuned model has difficulties utilizing this knowledge. The model also has difficulties understanding some paraphrased statements, e.g., from \textit{over 1/2 patients} to \textit{over 0.5 patients}, from \textit{5\% of patients} to \textit{0.05 of patients}, etc. Another issue is some conversions are not quite precise, e.g., \textit{9 months} to \textit{36 weeks}, \textit{6 months} to \textit{180 days}. The model often contradicts the modified statements since they are not the same (9 months equals about 39 weeks).

\begin{table}[t]
\centering
\resizebox{\columnwidth}{!}{
\begin{tabular}{@{}lccccc@{}}
\toprule
\multicolumn{1}{c}{}&\multicolumn{5}{c}{\textbf{Consistency}}\\
\cline{2-6}
\textbf{Model} & \textbf{para.} &\textbf{cont.}  &\textbf{num. para.}  &\textbf{num. cont.} &\textbf{defi.}\\
\midrule
Label-only Greedy   &0.7807           &0.7613          &\textbf{0.7411}               &0.9568              &\textbf{0.6553}       \\
CoT Greedy    &0.7780            &0.8427          &0.7321               &0.9568              &0.5780        \\
Self-consistent CoT& \textbf{0.8020}     &\textbf{0.8440}          &0.7232               &\textbf{0.9877}              &0.5720        \\

\bottomrule
\end{tabular}%
}
\caption{Consistency scores of three systems according to semantic-preserving intervention types}
\label{tab:consistency_comparison}
\end{table}

\subsection{Faithfulness}
The contrast set for evaluating the faithfulness of the system is generated from 250 samples from the control set. All the 250 statements in the control set have the label Entailment. With contradiction intervention, 864 samples are created with the label Contradiction, 114 of which are numerically intervened. \Cref{tab:faithfulness_comparison} summarizes the faithfulness scores of each system after contradiction intervention types. The scores show that both CoT greedy and self-consistent CoT have a significant performance improvement over label-only prediction. This underscores the critical role of the extra reasoning chain generation step for faithful verification.

\subsection{Self-consistent CoT and CoT Greedy}
According to Table~\ref{tab:overall_comparison}, our self-consistent CoT has improved the performance of the CoT greedy system regarding all metrics, namely baseline F1 of 1.31 percentage points, consistency score of 0.75 percentage points, and faithfulness score of 0.69 percentage points. However, the improvement is insignificant compared to \citet{wang2023selfconsistency}. We show two examples in Appendix~\ref{appendix:generation}. The first example shows that self-consistent CoT corrects the error in CoT greedy, and the second one shows that self-consistent CoT fails to correct the error. One possible reason for the insignificant improvement is the number of generated reasoning chains. Due to computational limitations, we have only generated 10.36 reasoning chains on average, which is much less than the 40 reasoning chains in \citet{wang2023selfconsistency}. Another important reason is the aggregating mechanism of majority voting. From the above-mentioned second example, we can see the late simple aggregation of the verification labels is not enough for the binary classification problem. A more finegrained verification and integration of intermediate reasoning steps of CoT is needed to tackle the inconsistency problem further.

\begin{table}[t]
\centering
\resizebox{\columnwidth}{!}{
\begin{tabular}{@{}lcc@{}}
\toprule
\multicolumn{1}{c}{}&\multicolumn{2}{c}{\textbf{Faithfulness}}\\
\cline{2-3}
\textbf{Model}                  &\textbf{cont. alter.}  &\textbf{num. cont. alter.}  \\
\midrule
Label-only Greedy               &0.8040                  &0.8509                      \\
CoT Greedy                      &0.8933                  &0.9211                      \\
Self-consistent CoT             &\textbf{0.8987}                  &\textbf{0.9386}                      \\
\bottomrule
\end{tabular}%
}
\caption{Faithfulness scores of three systems according to intervention (altering) types}
\label{tab:faithfulness_comparison}
\end{table}

\section{Conclusion}
In this paper, we have reported our self-consistent CoT system for the SemEval-2024 Task 2: Safe Biomedical Natural Language Inference for Clinical Trials. For comparison, we also reported the label-only greedy and CoT greedy inference systems. To achieve a trustworthy inference system, we utilized the CoT reasoning paradigm, not only predicting the verification labels but also rationales. We tackled CoT's inconsistency problem with self-consistent CoT. Compared to the greedy CoT system, we have improved the inference performance by generating multiple reasoning chains and verifying with majority voting. However, the performance improvement is limited. For future work, a more fine-grained evaluation of the correctness of the reasoning steps in the CoT paradigm is promising for solving complex reasoning tasks.

\section{Acknowledgments}
This work was carried out with the support of the German Federal Ministry of Education and Research (BMBF) within the project "DeFaktS" (Grant 16KIS1524K). This work was also supported by the Helmholtz Association Initiative and Networking Fund on the HAICORE@KIT partition.

\bibliography{acl_latex}
\newpage
\appendix

\section{Prompts}
\subsection{Instruction prompt for GPT-4}
\Cref{fig:gpt4_example} shows an instruction example for GPT-4 to verify the statements in the train and validation datasets with CoT.
\label{sec:instruction_gpt4}
\begin{figure}[ht]
\noindent\fbox{\begin{minipage}{19em}
\footnotesize{
<s>[INST] \textbf{Primary clinical trial report}:

Adverse Events 1: Total: 12/32 (37.50\%) Anaemia 0/32 (0.00\%) Neutropenia 1/32 (3.13\%) Thrombocytopenia 4/32 (12.50\%) Atrial fibrillation 1/32 (3.13\%) Cardiac failure congestive 1/32 (3.13\%) Myocardial ischaemia 1/32 (3.13\%) Abdominal discomfort 0/32 (0.00\%) Ascites 1/32 (3.13\%) Constipation 0/32 (0.00\%) Rectal haemorrhage 1/32 (3.13\%) Vomiting 1/32 (3.13\%) Fatigue 1/32 (3.13\%) Adverse Events 2: Total: 8/20 (40.00\%) Anaemia 1/20 (5.00\%) Neutropenia 0/20 (0.00\%) Thrombocytopenia 1/20 (5.00\%) Atrial fibrillation 0/20 (0.00\%) Cardiac failure congestive 0/20 (0.00\%) Myocardial ischaemia 0/20 (0.00\%) Abdominal discomfort 1/20 (5.00\%) Ascites 0/20 (0.00\%) Constipation 2/20 (10.00\%) Rectal haemorrhage 0/20 (0.00\%) Vomiting 0/20 (0.00\%) Fatigue 0/20 (0.00\%)

\textbf{Statement}: Most of the cases of CHF in the primary trial, were in cohort 1.

Verify whether the statement is entailed in the primary clinical trial report with Entailment or Contradiction. [/INST]

\textbf{Let's verify it step by step}:
}
\end{minipage}}
\caption{Example prompt for GPT-4}
\label{fig:gpt4_example}
\end{figure}

\subsection{Example for CoT instruction-tuning dataset}
\label{sec:cot_instrcution_data_example}
\Cref{fig:cot_instruction_dataset_example} shows an example of CoT instruction-tuning dataset. The rationales in CoT are generated by GPT-4 and we add the final verification label to CoT for convenience by extracting the labels.
\begin{figure}[ht]
\noindent\fbox{\begin{minipage}{19em}
\footnotesize{
<s>[INST] \textbf{Primary clinical trial report}:

Adverse Events 1: Total: 12/32 (37.50\%) Anaemia 0/32 (0.00\%) Neutropenia 1/32 (3.13\%) Thrombocytopenia 4/32 (12.50\%) Atrial fibrillation 1/32 (3.13\%) Cardiac failure congestive 1/32 (3.13\%) Myocardial ischaemia 1/32 (3.13\%) Abdominal discomfort 0/32 (0.00\%) Ascites 1/32 (3.13\%) Constipation 0/32 (0.00\%) Rectal haemorrhage 1/32 (3.13\%) Vomiting 1/32 (3.13\%) Fatigue 1/32 (3.13\%) Adverse Events 2: Total: 8/20 (40.00\%) Anaemia 1/20 (5.00\%) Neutropenia 0/20 (0.00\%) Thrombocytopenia 1/20 (5.00\%) Atrial fibrillation 0/20 (0.00\%) Cardiac failure congestive 0/20 (0.00\%) Myocardial ischaemia 0/20 (0.00\%) Abdominal discomfort 1/20 (5.00\%) Ascites 0/20 (0.00\%) Constipation 2/20 (10.00\%) Rectal haemorrhage 0/20 (0.00\%) Vomiting 0/20 (0.00\%) Fatigue 0/20 (0.00\%)

\textbf{Statement}: Most of the cases of CHF in the primary trial, were in cohort 1.

Verify whether the statement is entailed in the primary clinical trial report with Entailment or Contradiction. [/INST]

\textbf{Let's verify it step by step}: The statement refers to cases of Cardiac Heart Failure (CHF) in the primary trial. Looking at the data from the primary clinical trial report, we can see that in Adverse Events 1 (cohort 1), there was 1 case of Cardiac failure congestive out of 32 participants, which is 3.13\%. In Adverse Events 2 (cohort 2), there were 0 cases of Cardiac failure congestive out of 20 participants, which is 0.00\%.

Therefore, the statement "Most of the cases of CHF in the primary trial, were in cohort 1" is Entailed in the primary clinical trial report.

\textbf{Verification}: \textcolor{green}{Entailment} </s>
}
\end{minipage}}
\caption{An example of CoT instruction-tuning dataset}
\label{fig:cot_instruction_dataset_example}
\end{figure}

\subsection{Example for label-only instruction-tuning dataset}
\label{sec:label_only_instrcution_data_example}
\Cref{fig:label_only_instruction_dataset_example} shows an example of the label-only instruction-tuning dataset, where only the gold label is appended to the (premise, statement) pair without CoT rationales.
\begin{figure}[ht]
\noindent\fbox{\begin{minipage}{19em}
\footnotesize{
<s>[INST] \textbf{Primary clinical trial report}:

Adverse Events 1: Total: 12/32 (37.50\%) Anaemia 0/32 (0.00\%) Neutropenia 1/32 (3.13\%) Thrombocytopenia 4/32 (12.50\%) Atrial fibrillation 1/32 (3.13\%) Cardiac failure congestive 1/32 (3.13\%) Myocardial ischaemia 1/32 (3.13\%) Abdominal discomfort 0/32 (0.00\%) Ascites 1/32 (3.13\%) Constipation 0/32 (0.00\%) Rectal haemorrhage 1/32 (3.13\%) Vomiting 1/32 (3.13\%) Fatigue 1/32 (3.13\%) Adverse Events 2: Total: 8/20 (40.00\%) Anaemia 1/20 (5.00\%) Neutropenia 0/20 (0.00\%) Thrombocytopenia 1/20 (5.00\%) Atrial fibrillation 0/20 (0.00\%) Cardiac failure congestive 0/20 (0.00\%) Myocardial ischaemia 0/20 (0.00\%) Abdominal discomfort 1/20 (5.00\%) Ascites 0/20 (0.00\%) Constipation 2/20 (10.00\%) Rectal haemorrhage 0/20 (0.00\%) Vomiting 0/20 (0.00\%) Fatigue 0/20 (0.00\%)

\textbf{Statement}: Most of the cases of CHF in the primary trial, were in cohort 1.

Verify whether the statement is entailed in the primary clinical trial report with Entailment or Contradiction. [/INST]

\textbf{Verification}: \textcolor{green}{Entailment} </s>
}
\end{minipage}}
\caption{An example of label-only instruction-tuning dataset}
\label{fig:label_only_instruction_dataset_example}
\end{figure}

\section{Implementation Details}
\subsection{Instruction-Tuning}
\label{appendix:instruction_tuning}
For the instruction-tuning, we have applied PyTorch Fully Sharded Data Parallel (FSDP) from Meta AI \cite{pytorch_fsdp}. The training script is adapted based on llama-recipes\footnote{\url{https://github.com/facebookresearch/llama-recipes/tree/main}}. We train Mixtral-8x7B-Instruct on a cluster with 16 A100-40GB GPUs with a batch size of 1 and a learning rate of 0.0001. Mixed precision and gradient checkpointing have been applied to save memory. Evaluated on the validation dataset, the model achieves the best performance with 5 epochs. Following \citet{kanakarajan-sankarasubbu-2023-saama}, we include the validation set as part of the training data for the final submission.

\subsection{Model Inference}
\label{appendix:model_inference}
We use the transformers\footnote{\url{https://github.com/huggingface/transformers}} library to generate responses with instruction-tuned models. There are two setups for generation, greedy decoding and temperature sampling. Under temperature sampling, we set $T=0.7$ and $k=50$ in top-$k$. For each pair, we set \texttt{num\_return\_sequence} as 10. Occasionally, there are duplicates in the returned sequences. If the result is tied with majority voting, we further generate reasoning chains with different seeds. The maximum number of generated chains for one pair is 25. \Cref{tab:distribution_chains} shows the distribution for numbers of generated distinct reasoning chains in the test dataset. On average, we generate 10.36 reasoning chains for each pair in the test dataset.

\begin{table}[t]
\centering
\resizebox{0.85\columnwidth}{!}{
\begin{tabular}{@{}rr@{}}
\toprule
\textbf{Number of Chains} & \textbf{Number of Samples} \\
\midrule
10 & 5150\\
9 & 104 \\
15 &101\\
20 &90\\
25&36\\
19&8\\
17&5\\
16&2\\
8, 14, 18, 24 &1 \\ 
\bottomrule
\end{tabular}%
}
\caption{Distribution of numbers of generated distinct reasoning chains for test dataset }
\label{tab:distribution_chains}
\end{table}


%
%

\section{Reasoning Chains Generation}
Figure~\ref{fig:successful_correction} shows an example where CoT greedy has wrong verification while self-consistent CoT has correct verification with majority voting. Figure~\ref{fig:failed_correction} shows an example where both CoT greedy and self-consistent CoT have wrong verification.
\label{appendix:generation}
\begin{figure*}[ht]
\noindent\fbox{\begin{minipage}{40em}
\footnotesize{
<s>[INST] \textbf{Primary clinical trial report}:
INTERVENTION 1: Arm 1 Herceptin/navelbine INTERVENTION 2: Arm 2 Taxotere/carboplatin/herceptin

\textbf{Statement}: only one arm of the primary clinical trial makes use of trastuzumab as part of its intervention

Verify whether the statement is entailed in the primary clinical trial report with Entailment or Contradiction. [/INST]

Let's verify it step by step:\\

\textbf{Gold Label}: \textcolor{red}{Contradiction}\\

\textbf{CoT Greedy}:

The statement mentions that only one arm of the primary clinical trial makes use of trastuzumab as part of its intervention. In the clinical trial report, Herceptin is mentioned in both Intervention 1 and Intervention 2. Herceptin is a brand name for trastuzumab, a monoclonal antibody used to treat certain types of breast cancer. Therefore, the statement is Entailed in the primary clinical trial report.

Verification: \textcolor{green}{Entailment} \\

\textbf{Self-consistent CoT}

Generation 1:

The statement mentions that only one arm of the primary clinical trial makes use of trastuzumab as part of its intervention. Trastuzumab is a monoclonal antibody used to treat certain types of breast cancer. It is also known by its brand name, Herceptin. In the primary clinical trial report, Herceptin is mentioned in both Intervention 1 (Arm 1) and Intervention 2 (Arm 2). Therefore, both arms of the trial are using trastuzumab (Herceptin) in their interventions. So, the statement is a Contradiction to the information provided in the primary clinical trial report.

Verification: \textcolor{red}{Contradiction} 

Generation 2: 

The statement mentions that only one arm of the primary clinical trial makes use of trastuzumab as part of its intervention. Looking at the clinical trial report, we can see that Herceptin is used in both arms of the trial. Herceptin is a brand name for trastuzumab, a medication used to treat certain types of breast cancer. Therefore, the statement is a Contradiction because both arms of the trial use trastuzumab, not just one.

Verification: \textcolor{red}{Contradiction}

Generation 3:

The statement mentions that only one arm of the primary clinical trial makes use of trastuzumab as part of its intervention. Trastuzumab is the generic name for Herceptin. Looking at the clinical trial report, we see that Herceptin is indeed used in both interventions. In Intervention 1, Herceptin is used in combination with navelbine. In Intervention 2, Herceptin is used in combination with Taxotere and carboplatin. Therefore, the statement contradicts the information given in the primary clinical trial report. The correct answer is Contradiction.

Verification: \textcolor{red}{Contradiction} 

Generation 4:

The statement mentions that only one arm of the primary clinical trial makes use of trastuzumab as part of its intervention. In the clinical trial report, Herceptin is mentioned in both Intervention 1 and Intervention 2. Herceptin is a brand name for trastuzumab, a monoclonal antibody used to treat certain types of breast cancer. Therefore, the statement is Entailed in the primary clinical trial report.

Verification: \textcolor{green}{Entailment}

Generation 5:

The clinical trial report mentions two interventions. In Intervention 1 (Arm 1), Herceptin is mentioned. Herceptin is a brand name for the drug trastuzumab. In Intervention 2 (Arm 2), there is no mention of Herceptin or trastuzumab. Therefore, the statement "only one arm of the primary clinical trial makes use of trastuzumab as part of its intervention" is Entailed in the primary clinical trial report.
Verification: \textcolor{green}{Entailment}

{......}

}
\end{minipage}}
\caption{An example of self-consistent CoT successfully corrects the error in CoT greedy verification. We only show 5 generations of self-consistent CoT. The ratio of 10 self-consistent CoT generations is 8 Contradictions: 2 Entailments. }
\label{fig:successful_correction}
\end{figure*}

\begin{figure*}[ht]
\noindent\fbox{\begin{minipage}{40em}
\footnotesize{
<s>[INST] \textbf{Primary clinical trial report}:
Adverse Events 1: Total: 10/30 (33.33\%) Hemoglobin decreased 2/30 (6.67\%) Abdominal pain 1/30 (3.33\%) Colitis 1/30 (3.33\%) Diarrhea 7/30 (23.33\%) Nausea 2/30 (6.67\%) Rectal hemorrhage 1/30 (3.33\%) Fatigue 1/30 (3.33\%) Skin infection 1/30 (3.33\%) Neutrophil count decreased 1/30 (3.33\%) Platelet count decreased 3/30 (10.00\%) Dehydration 1/30 (3.33\%)

\textbf{Secondary clinical trial report}:
Adverse Events 1: Total: 3/6 (50.00\%) Anaemia 0/6 (0.00\%) Febrile neutropenia 0/6 (0.00\%) Neutropenia 0/6 (0.00\%) Bradycardia 0/6 (0.00\%) Diarrhoea 0/6 (0.00\%) Pancreatitis 0/6 (0.00\%) Vomiting 0/6 (0.00\%) Disease progression 0/6 (0.00\%) Fatigue 0/6 (0.00\%) Pyrexia 0/6 (0.00\%) Cholelithiasis 0/6 (0.00\%) Hepatic pain 0/6 (0.00\%) Bacteraemia 0/6 (0.00\%) Adverse Events 2: Total: 3/6 (50.00\%) Anaemia 0/6 (0.00\%) Febrile neutropenia 0/6 (0.00\%) Neutropenia 1/6 (16.67\%) Bradycardia 0/6 (0.00\%) Diarrhoea 0/6 (0.00\%) Pancreatitis 0/6 (0.00\%) Vomiting 0/6 (0.00\%) Disease progression 0/6 (0.00\%) Fatigue 1/6 (16.67\%) Pyrexia 1/6 (16.67\%) Cholelithiasis 0/6 (0.00\%) Hepatic pain 1/6 (16.67\%) Bacteraemia 0/6 (0.00\%)

\textbf{Statement}: diarrhoea was uncommon among the primary trial participants compared to the secondary trial participants

Verify whether the statement is entailed in the clinical trial reports with Entailment or Contradiction. [/INST]

Let's verify it step by step:\\

\textbf{Gold Label}: \textcolor{red}{Contradiction}\\

\textbf{CoT Greedy}:

In the primary clinical trial report, the occurrence of diarrhea is 7/30 (23.33\%). 
In the secondary clinical trial report, the occurrence of diarrhea is 0/6 (0.00\%) in both Adverse Events 1 and Adverse Events 2. Therefore, the statement "diarrhoea was uncommon among the primary trial participants compared to the secondary trial participants" is Entailed in the clinical trial reports.

Verification: \textcolor{green}{Entailment} \\

\textbf{Self-consistent CoT}

Generation 1:

The statement claims that diarrhoea was uncommon among the primary trial participants compared to the secondary trial participants. Looking at the primary clinical trial report, it shows that the incidence of diarrhoea was 7 out of 30 participants, which is approximately 23.33\%. In the secondary clinical trial report, the incidence of diarrhoea is 0 out of 6 participants in both Adverse Events 1 and Adverse Events 2, which is approximately 0.00\%. Comparing these two reports, it is clear that diarrhoea was indeed more common in the primary trial participants than in the secondary trial participants. Therefore, the statement is a Contradiction to the information provided in the clinical trial reports.

Verification: \textcolor{red}{Contradiction} 

Generation 2: 

The statement says that diarrhoea was uncommon among the primary trial participants compared to the secondary trial participants. Looking at the primary clinical trial report, Diarrhea was reported in 7 out of 30 participants, which is approximately 23.33\%. In the secondary clinical trial report, Diarrhoea was reported in 0 out of 6 participants in both Adverse Events 1 and Adverse Events 2, which is 0\%. Comparing these two, it is clear that Diarrhoea was indeed more common in the primary trial participants than in the secondary trial participants. Therefore, the statement is a Contradiction to the information provided in the clinical trial reports.

Verification: \textcolor{red}{Contradiction}

Generation 3:

In the primary clinical trial report, the incidence of diarrhea was 7/30 (23.33\%). In the secondary clinical trial report, the incidence of diarrhoea was 0/6 (0.00\%) for both Adverse Events 1 and Adverse Events 2. Therefore, the statement "diarrhoea was uncommon among the primary trial participants compared to the secondary trial participants" is Entailed in the clinical trial reports.

Verification: \textcolor{green}{Entailment} 

Generation 4:

In the primary clinical trial report, the occurrence of diarrhea is reported as 7/30 (23.33\%). In the secondary clinical trial report, the occurrence of diarrhea is reported as 0/6 (0.00\%) in both Adverse Events 1 and Adverse Events 2. Therefore, the statement "diarrhoea was uncommon among the primary trial participants compared to the secondary trial participants" is Entailed in the clinical trial reports.

Verification: \textcolor{green}{Entailment}

Generation 5:

In the primary clinical trial report, Diarrhea is reported in 7 out of 30 participants, which is approximately 23.33\%. In the secondary clinical trial report, Diarrhoea is reported in 0 out of 6 participants in both Adverse Events 1 and Adverse Events 2, which is 0.00\%. Therefore, the statement "diarrhoea was uncommon among the primary trial participants compared to the secondary trial participants" is Entailed in the clinical trial reports.

Verification: \textcolor{green}{Entailment}

{......}

}
\end{minipage}}
\caption{An example of self-consistent CoT fails to correct the error in CoT greedy verification. We only show 5 generations of self-consistent CoT. The ratio of 10 self-consistent CoT generations is 7 Entailments: 3 Contradictions. }
\label{fig:failed_correction}
\end{figure*}

\end{document}